\documentclass[conference]{IEEEtran}
\IEEEoverridecommandlockouts

\usepackage{multirow}

\usepackage{cite}

\usepackage{amssymb,amsfonts,amsthm}
\usepackage[fleqn]{amsmath}

\usepackage{algorithmic}
\usepackage{graphicx}
\usepackage{textcomp}
\usepackage{xcolor}

\usepackage{algorithm}

\usepackage{graphicx}
\usepackage{caption}
\usepackage{subcaption}
\usepackage{url}

\usepackage{balance}

\usepackage{mdframed}

\def\BibTeX{{\rm B\kern-.05em{\sc i\kern-.025em b}\kern-.08em
    T\kern-.1667em\lower.7ex\hbox{E}\kern-.125emX}}

\usepackage{todonotes}

\DeclareCaptionLabelFormat{andtable}{#1~#2  \&  \tablename~\thetable}

\begin{document}

\title{Leveraging Trust and Distrust in Recommender Systems via Deep Learning}

\author{\IEEEauthorblockN{Dimitrios Rafailidis}
\IEEEauthorblockA{\textit{Maastricht University} \\
Maastricht, The Netherlands \\
dimitrios.rafailidis@maastrichtuniversity.nl}
}

\maketitle

\begin{abstract}
The data scarcity of user preferences and the cold-start problem often appear in real-world applications and limit the recommendation accuracy of collaborative filtering strategies. Leveraging the selections of social friends and foes can efficiently face both problems.  In this study, we propose a strategy that performs social deep pairwise learning. Firstly, we design a ranking loss function incorporating multiple ranking criteria based on the choice in users, and the choice in their friends and foes to improve the accuracy in the top-k recommendation task. We capture the nonlinear correlations between user preferences and the social information of trust and distrust relationships via a deep learning strategy. In each backpropagation step, we follow a social negative sampling strategy to meet the multiple ranking criteria of our ranking loss function. We conduct comprehensive experiments on a benchmark dataset from Epinions, among the largest publicly available that has been reported in the relevant literature. The experimental results demonstrate that the proposed model beats other state-of-the art methods, attaining an 11.49\% average improvement over the most competitive model. We show that our deep learning strategy plays an important role in capturing the nonlinear correlations between user preferences and the social information of trust and distrust relationships, and demonstrate the importance of our social negative sampling strategy on the proposed model.
\end{abstract}


\section{Introduction} \label{sec:intro}

Collaborative filtering has been widely adopted in recommendation systems, providing similar-minded users with similar recommendations~\cite{p9,KorenBV09}. In the real-setting though, the data scarcity of user preferences and the cold-start problem degrade the quality of recommendations. To face these problems, various collaborative filtering strategies exploit the selections of social friends~\cite{TRUSTSVD,SOREC}. Real-world platforms such as Last.fm\footnote{\url{https://www.last.fm/}}, allow users either to form trust relationships or import friends from other platforms such as Facebook\footnote{\url{https://www.facebook.com/}} and Yahoo!\footnote{\url{https://yahoo.com/}}. Exploiting the selections of trust friends in collaborative filtering stategies can efficiently combat the data scarcity and cold-start problems as shown in~\cite{SOCIALMF,STE}. Moreover, users on Slashdot\footnote{\url{https://slashdot.org/}}, a technological-related news and discussion site, can directly tag other users as friends and foes. Epinions\footnote{\url{http://www.epinions.com/}}, an e-commerce site for reviewing and rating products, allows users to evaluate other users based on the quality of their reviews, and establish trust and distrust relations with them. In recognition that users might accept the recommendations from the friends but not from their foes, several models exploit both trust and distrust relationships in recommendation systems~\cite{FOR15,FOR14, MA09, TAN16,Rafailidis16a}. However, these studies consider the explicit feedback of ratings using pointwise loss functions in their learning strategies to produce recommendations.

In practice, rather than explicit feedbacks of ratings users' implicit feedback is usually available, such as number of views, clicks and so on. Several methods have been proposed to handle the case of users' implicit feedback, such as weighted matrix factorization~\cite{p9}. These methods design pointwise loss functions such as square or cross-entropy loss functions to predict either the error prediction or if an unobserved item would be preferred or not by a user. However, provided that end-users are usually interested in the top-$k$ recommendations, pointwise loss functions do not necessarily focus on the top-$k$ recommendation task~\cite{RecSys12}. For instance, if all the low-ranked ratings are predicted very accurately, but significant errors are made on the higher-ranked ratings, then these methods will provide a low-quality personalized recommendation list to users. 

In the context of pairwise learning~\cite{SemertzidisRSD15}, aiminig at the top-$k$ recommendation task Bayesian Personalized Ranking (BPR) strategies use a \emph{pairwise ranking loss} function, considering the relative ordering of items in a ranked list~\cite{BPR}. The pairwise ranking criterion of the BPR model is based on the assumption that a user prefers the observed items over the unobserved ones. This idea results in a pairwise ranking loss function that tries to discriminate between a small set of observed items and a very large set of unobserved ones. Due to the imbalance between the user's observed items and unobserved ones, the BPR model uniformly samples negative examples from the set of unobserved items to reduce the training time, also known as \emph{negative sampling}. In a similar spirit, other representative ranking models among recommendation systems are CofiRank~\cite{CofiRank} and CLiMF~\cite{CLIMF}, which use loss functions based on Normalized Discounted Cumulative Gain and Reciprocal Rank, respectively. However, the aforementioned studies ignore the social information of trust and distrust.

While several ranking models follow the pairwise learning strategy of BPR to exploit the selections of friends such as~\cite{p15,WangLEWC16,CIKM14}, they do not account for the selections of foes. More recently, in~\cite{Rec17} a ranking model with trust and distrust relationships is proposed based on the optimization ranking algorithm of~\cite{WWW15}. However, this model does not capture the nonlinear correlations between user preferences and the social information of trust and distrust relationships, thus having limited recommendation accuracy as we will experimentally show in Section~\ref{sec:exp}. Meanwhile, deep learning stategies have been proved to be an effective means to compute the nonlinear correlations between different types of user data in recommendation systems, due to their ability of producing abstract representations at the hidden layers of their deep structures~\cite{p29,p34,p26,p22,p23,p25}. Also, a few attempts have been made to consider user preferences and trust relationships in collaborative filtering with deep learning strategies~\cite{DengHXWW17,NguyenL16,RafailidisC17}. However, they ignore users' distrust relationships and consequently produce less accurate recommendations~\cite{MA09}.  

To overcome the shortcomings of existing methods we propose a social ranking model that performs deep pairwise learning with trust and distrust relationships, namely SDPL, making the following contributions:

\begin{itemize}
\item We formulate our ranking problem following the BPR framework. We design a ranking loss function with multiple ranking criteria based on the choice in users, and the choice in their friends and foes. 
\vspace{0.1cm}
\item We propose a deep learning strategy to optimize the ranking loss function with the multiple ranking criteria, to calculate the nonlinear correlations between user preferences and the social information of trust and distrust relationships at the deep representations. We learn the parameters of our ranking model via backpropagation. In this respect, in each backpropagation step we follow a social negative sampling strategy to meet the multiple ranking criteria of our ranking loss function.
\end{itemize}

\noindent Our experiments on a benchmark dataset from Epinions, among the largest publicly available, show that the proposed social deep pairwise learning model outperforms competitive models when the data scarcity of user preferences varies, as well as in the case of cold-start users. Also, we demonstrate the importance of our deep learning strategy to capture the nonlinear correlations between user preferences and users' social information of trust and distrust. Furthermore, we experimentally show that our social negative sampling strategy is a key factor for the proposed SDPL model.

The remainder of the paper is organized as follows, Section~\ref{sec:rel} reviews the related work and in Section~\ref{sec:prop} we present the proposed SDPL model, finally, Section~\ref{sec:exp} presents the experimental results and Section~\ref{sec:conc} concludes the study. 

\section{Related Work}\label{sec:rel}

\subsection{Social Recommendation}

\paragraph{Ranking models with trust} Instead of using pointwise loss functions such as~\cite{TRUSTSVD,SOCIALMF,STE,SOREC}, several ranking models with trust relationships have been widely studied in recommendation systems. For example, \cite{CIKM14} present a trust-based BPR model that incorporates trust relationships into a pairwise ranking model, assuming that users tend to assign higher ranks to items that their friends prefer. \cite{WangLEWC16} introduce a BPR model with strong and weak social ties of the user's friends. Their ranking criterion assumes that the observed items of weak tie friends should be ranked higher than those of strong tie friends, because weak tie friends are likely to introduce novel and diverse items. \cite{p15} incorporate friends' selections and the geographical proximity of venues in the negative sampling strategy of BPR to produce recommendations. Except for the BPR framework, other studies use different ranking models for the top-$k$ social recommendation task with trust. For instance,~\cite{RAF16b,RafailidisC16a} jointly learn different ranking strategies into a unified model to leverage the recommendation accuracy with trust relationships.~\cite{SPF} infer user preferences and the influence of her friends based on Social Poisson factorization. The main limitation of the above ranking models is that they ignore users' distrust relationships, a crucial factor to leverage the quality of recommendations~\cite{Rec17,TAN16}. Our proposed SDLP model accounts for both the selections of social friends and foes when learning the ranking model, thus significantly improving the quality of recommendations.

\paragraph{Models with trust and distrust} In~\cite{MA09} a trust-based, as well as a distrust-based model are introduced to exploit trust and distrust relationships in each model separately. The goal of the trust-based model is to minimize the distances of latent features between friends, while the distrust-based model tries to maximize the latent features' distances between foes. To exploit both trust and distrust relationships at the same time in a unified model,~\cite{FOR14} use trust and distrust relationships in a matrix factorization framework using a hinge loss function. This method assumes that the trust/distrust relationships between users are considered as similarity/dissimilarity in their preferences. Then, the latent features are computed in a manner such that the latent features of foes who are distrusted by a certain user have a guaranteed minimum dissimilarity gap from the worst dissimilarity of friends who are trusted by this same user. In~\cite{FOR15}, a recommendation strategy is introduced to rank users' latent features based on their trust and distrust relationships, considering also neutral relationships between users who have no relation to a certain user. The goal of this strategy is to rank the neutral users' latent features after the friends' latent features and before those of foes.~\cite{TAN16} form a signed graph where trust and distrust relationships are modeled into edges with positive and negative weights, respectively. This model first captures local and global information from the signed graph. Local information reveals the correlations among users and her friends/foes, and the global information reveals the reputation of the user in the whole social network, as users tend to trust users with high global reputation. Then, they exploit both local and global information in a matrix factorization technique to compute the recommendations. The main drawback of these methods is that they do not focus on the top-$k$ recommendation task, as they design different poinwise loss functions in their learning strategies. In~\cite{Rec17}, a ranking model is presented to balance the influences of friends' and foes' selections on users' preferences based on the collaborative ranking algorithm of~\cite{WWW15}, focusing on the accuracy at the top of the list. The main limitation of the ranking model of ~\cite{Rec17} is that it does not compute the nonlinear associations of user preferences and the social information of trust and distrust relationships. Instead, our deep learning model can capture the nonlinear associations of user preferences and those of their friends and foes,  outperforming several models with trust and distrust, as we will experimentally show in Section~\ref{sec:exp}.

\subsection{Deep Learning for Collaborative Filtering}

\paragraph{Neural collaborative filtering} Accordingly, in recommendation systems deep learning strategies use either a pointwise~\cite{p26} or a pairwise ranking loss function~\cite{p25} to handle user implicit feedbback and capture user data nonlinear correlations.  In~\cite{p29}, different sampling strategies are examined for implicit feedback to train neural network-based collaborative filtering strategies with pointwise or pairwise loss functions.  In~\cite{p22,p23,p21}, various deep learning strategies are introduced to exploit user feedback with users' and items' side information. For example,~\cite{p21} model implicit feedback in stacked Denoising Autoencoders with the side information of articles, such as the title and abstract of the articles. Compared to the proposed SDPL model, the above studies do not consider any social information when training their deep learning models, a key factor to generate accurate recommendations~\cite{TRUSTSVD,SOCIALMF,Rec17,TAN16}. 

\paragraph{Trust-based recommendation with deep learning} Recently, a few deep learning strategies have been introduced to generate recommendations with social relationships, such as Denoising Autoencoders~\cite{DengHXWW17} and Restricted Boltzmann Machines~\cite{NguyenL16}.~\cite{p1} introduce a deep neural model that jointly learns the embeddings of users and locations to predict user preferences over locations. This model first transforms the users' social relationships and locations' contextual proximities into graphs and then employs neural embedding for location recommendation as a bridge between collaborative filtering and semi-supervised learning. Instead of using a pairwise ranking function, it defines a pointwise function to handle the case of implicit feedback during the deep neural network learning. All the aforementioned studies are limited to users' trust relationships and do not account for the selections of foes. Likewise,~\cite{p34} extend the BPR model by first extracting deep features based on a convolutional neural network, and then using a deep neural network to produce friend recommendations. Compared to the proposed SDPL model, trust-based recommendation with deep learning strategies do not account for both user preferences and distrust relationships, thus having limited accuracy.

\subsection{Deep Network Embedding with Trust and Distrust Relationships}
Meanwhile, several deep network embedding strategies have been proposed in the literature, which aim to learn low-dimensional vector representations of users/nodes based on users' trust relationships. For example, DeepWalk~\cite{p36} introduces the idea of Skip-gram, a word representation model in NLP, to learn node representations from random-walk sequences.~\cite{p37} introduce the node2vec  model, to learn a mapping of nodes to a low-dimensional space of features that maximizes the likelihood of preserving network neighborhoods of nodes. They define a flexible notion of a node's network neighborhood and design a biased random walk procedure, which efficiently explores diverse neighborhoods. Similarly, various deep network embedding strategies have been introduced for signed graphs with both trust and distrust relationships~\cite{Wang17,WangATL17}. These embedding strategies generate the low-dimensional vector representations based on the structural balance theory~\cite{Cyg15}, \emph{``a structure in a signed graph should ensure that users should be able to have their friends closer than their foes''}. Instead of focusing on the recommendation task, the deep network embedding strategies omit user preferences,  aiming at different tasks such as link prediction, community detection and node classification.

\section{The Proposed Model} \label{sec:prop}

In our model we denote $\mathcal{N}$ and $\mathcal{I}$ the sets of users and items, with $n=|\mathcal{N}|$ being the number of users and $m=|\mathcal{I}|$ the number of items. The input of our model is a user-item matrix $X\in \mathbb{R}^{n \times m}$ with user preferences, expressed by any type of user-item interactions such as the explicit feedback of ratings or user implicit feedback such as number of views, clicks, and so on. In addition, we consider two adjacency matrices $A^+\in \mathbb{R}^{n \times n}$ and $A^-\in \mathbb{R}^{n \times n}$, corresponding to users' trust and distrust relationships. For example, if users $u$ and $a$ are friends then $A^+_{ua}=1$, and $A^+_{ua}=0$, otherwise. Likewise, if users $u$ and $b$ are foes then $A^-_{ub}=1$. For each user $u$ we calculate the neighborhoods $\mathcal{N}^+_u$ and $\mathcal{N}^-_u$ with her friends and foes based on the adjacency matrices $A^+$ and $A^-$.  

In the following, we first formulate the pairwise ranking problem (Section~\ref{sec:prob}) and design the ranking loss function of our model based on the multiple ranking criteria with trust and distrust relationships (Section~\ref{sec:loss}). Then, we present our deep pairwise learning architecture (Section~\ref{sec:arch}) and detail the model's parameter set $\mathbf{\Theta}$ in the hidden layers (Section~\ref{sec:hid}). Finally, we explain how to train our model via backpropagation based on a social negative sampling strategy (Section~\ref{sec:Train}).

\subsection{The Pairwise Ranking Problem} \label{sec:prob}

Given user preferences in $X$, users' trust relationships in $A^+$ and distrust ones in $A^-$, the goal of our model is to generate top-$k$ recommendations for a user $u \in {\mathcal{N}}$. In our SDPL model we formulate the recommendation problem as a pairwise ranking task~\cite{BPR}. We define the probability $x_{ui}$, where $x_{ui}=X(u,i)$ denotes that user $u$ has already interacted with item $i$. Thus, we can define two disjoint sets, a set $\mathcal{I}^+_u$ of observed items that user $u$ has already interacted with, and a set $\mathcal{I}^-_u$ of  unobserved items\footnote{In practice, we select users' unobserved items based on a social negative sampling strategy (Section~\ref{sec:Train}).}. For the task of top-$k$ recommendation, we build a pairwise ranking model that is able to rank the observed items before the unobserved ones. For any pair of items $i$ and $j$, with $i\in \mathcal{I}^+_u$ and $j\in \mathcal{I}^-_u$, the probability $x_{ui}$ should be greater than $x_{uj}$. To describe this relation we define a \emph{partial relation} $i >_u j$.   For each user $u\in \mathcal{N}$ the set of all partial relationships is computed as follows: 

\begin{equation} \label{eq:partial}
\mathcal{R}_u = \{i >_u j | i \in \mathcal{I}^+_u, j \in \mathcal{I}^-_u\} 
\end{equation}
We define our top-$k$ recommendation task of the proposed SDPL model as the following ranking problem:

\emph{\textbf{The Pairwise Ranking Problem:}}
``Given (i) the set of all partial relationships $\mathcal{R}_u$ for each user $ u\in \mathcal{N}$ (ii) users' trust relationships in $A^+$  and (iii) distrust ones in $A^-$, the goal of the proposed SDPL model is to maximize the ranking likelihood probability as follows:''
\begin{equation}
\max \prod_{u\in{\mathcal{N}}} \prod_{(i,j) \in \mathcal{R}_u} P(i >_u j)
\label{eq:lhood}
\end{equation}

\subsection{Ranking Loss Function} \label{sec:loss}
To account for the selections of friends and foes in our pairwise ranking problem of Eq.~(\ref{eq:lhood}), we consider the sets $\mathcal{I}^+_a$ and $\mathcal{I}^+_b$ which contain the observed items of a friend $a$ and foe $b$, with $a \in \mathcal{N}^+_u$ and  $b \in \mathcal{N}^-_u$. In particular, to design our ranking loss function we study the pairwise ordering of the observed and unobserved items of each user $u$, as well as the ordering of the selections of her friends and foes in the recommendation list. According to relevant studies, friends' preferences do not necessarily match, which means that the selections of user $u$ (observed items) have to be ranked higher than those of a friend $a$~\cite{CIKM14,WangLEWC16,SPF}.  In addition, foes' preferences certainly do not match as shown in the analysis of~\cite{MA09,FOR15,TAN16}. This can be interpreted as follows, the selections of user $u$ have to be ranked higher than those of a foe $b$, as well as the selections of a friend $a$ have to be ranked higher than those of a foe $b$. Hence, we reformulate the set $\mathcal{R}_u$ of all partial relationships of Eq.~(\ref{eq:partial}) based the following six cases of pairwise ordering $i >_u j$: 

\begin{equation}\label{eq:cases1}
i >_u j,\text{ if }\begin{cases}
\mathcal{R}_u^{(1)} = \{ i \in \mathcal{I}^+_u  \wedge j \in \mathcal{I}^-_u \wedge j  \notin \mathcal{I}^+_b  \} \\
\mathcal{R}_u^{(2)}=\{ i \in \mathcal{I}^+_u  \wedge j \in \mathcal{I}^+_a  \wedge j \notin  \mathcal{I}^+_u\}  \\
\mathcal{R}_u^{(3)}=\{ i \in \mathcal{I}^+_u  \wedge  i \in \mathcal{I}^+_a \wedge  j \in \mathcal{I}^+_a \}  \\
\mathcal{R}_u^{(4)}=\{ i \in \mathcal{I}^+_a  \wedge j \in \mathcal{I}^-_u \wedge j \notin \mathcal{I}^+_b \}   \\
\mathcal{R}_u^{(5)}=\{ i \in \mathcal{I}^+_a  \wedge j \in \mathcal{I}^+_b \} \\
\mathcal{R}_u^{(6)}=\{ i \in \mathcal{I}^-_u  \wedge j \in \mathcal{I}^+_b \} \\
\end{cases}
\end{equation}
The six cases of Eq.~(\ref{eq:cases1}) correspond to the following different ranking criteria:

\vspace{0.1cm}

\textit{(Case 1 - $\mathcal{R}_u^{(1)}$):} An observed item $i\in \mathcal{I}^+_u$ of user $u$ is ranked higher than an unobserved  item $j\in \mathcal{I}^-_u$, on condition that $j$  does not belong to the set $\mathcal{I}^+_b$  of observed items of foe $b$.

\vspace{0.1cm}

\textit{(Case 2 - $\mathcal{R}_u^{(2)}$):}  An observed item $i\in \mathcal{I}^+_u$ of user $u$ is ranked higher than an observed item $j \in \mathcal{I}^+_u$ of friend $a$. As there is a chance of user $u$ and friend $j$ to have partially interacted with common items $\mathcal{I}^+_u \cap \mathcal{I}^+_a \neq \emptyset$, in Case 2 we set $j \notin  \mathcal{I}^+_u$. The case of common items is covered by the Case 3 - $\mathcal{R}_u^{(3)}$.

\vspace{0.1cm}

\textit{(Case 3 - $\mathcal{R}_u^{(3)}$):}  A common observed item $i \in \mathcal{I}^+_u$ and $i \in \mathcal{I}^+_a$  of both user $u$ and friend $a$ is ranked higher than an item $j \in  \mathcal{I}^+_a$, where $j$ is an observed item only of friend $a$.

\vspace{0.1cm}

\textit{(Case 4 - $\mathcal{R}_u^{(4)}$):}  An observed item $i \in \mathcal{I}^+_a$ of friend $a$ is ranked higher than an unobserved item $j \in  \mathcal{I}^-_u$ of user $u$, on condition that $j$ is not in the set $\mathcal{I}^+_b $ of foe $b$.

\vspace{0.1cm}

\textit{(Case 5 - $\mathcal{R}_u^{(5)}$):}  An observed item $i\in \mathcal{I}^+_a $ of friend $a$ is ranked higher than an observed item $j\in \mathcal{I}^+_b $ of foe $b$.

\vspace{0.1cm}

\textit{(Case 6 - $\mathcal{R}_u^{(6)}$):}  An unobserved item $i\in \mathcal{I}^-_u $ of user $u$ is ranked higher than an observed item $j\in \mathcal{I}^+_b $ of foe $b$.

\vspace{0.1cm}

\noindent According to the $r=6$ ranking criteria of Eq.~(\ref{eq:cases1}) the pairwise ranking problem of Eq.~(\ref{eq:lhood}) is reformulated as follows:

\begin{equation}
\max \prod_{u\in{\mathcal{N}}} \prod_{r=1}^{6} \prod_{(i,j) \in \mathcal{R}_u^{(r)}} P(i >_u j)
\label{eq:lhood2}
\end{equation}

\noindent Using the negative natural logarithmic function $-\ln(x)$ and the sigmoid function $\sigma(x)=1/(1+e^{-x})$, we transform the maximization problem of Eq.~(\ref{eq:lhood2}) into a minimization problem, and we define our model's ranking loss function as follows:

\begin{align}  
\begin{aligned}
\min_{\mathbf{\Theta}}L &= - \sum_{u\in{\mathcal{N}}}  \sum_{r=1}^{6} \sum_{(i,j) \in \mathcal{R}_u^{(r)}} P(i >_u j) + \lambda || \mathbf{\Theta} ||^2 \\
 &= - \sum_{u\in{\mathcal{N}}}  \sum_{r=1}^{6} \sum_{(i,j) \in \mathcal{R}_u^{(r)}} \ln(\sigma(x_{ui}>x_{uj})) + \lambda || \mathbf{\Theta} ||^2 \\  
\label{eq:loss}
\end{aligned}
\end{align}
where $\mathbf{\Theta}$ is the model's parameter set (Section~\ref{sec:hid}) and $\lambda$ controls the regularization.

\subsection{Architecture} \label{sec:arch}

\begin{figure}[t]\centering
\includegraphics[width=8.5cm, height=6.6cm]{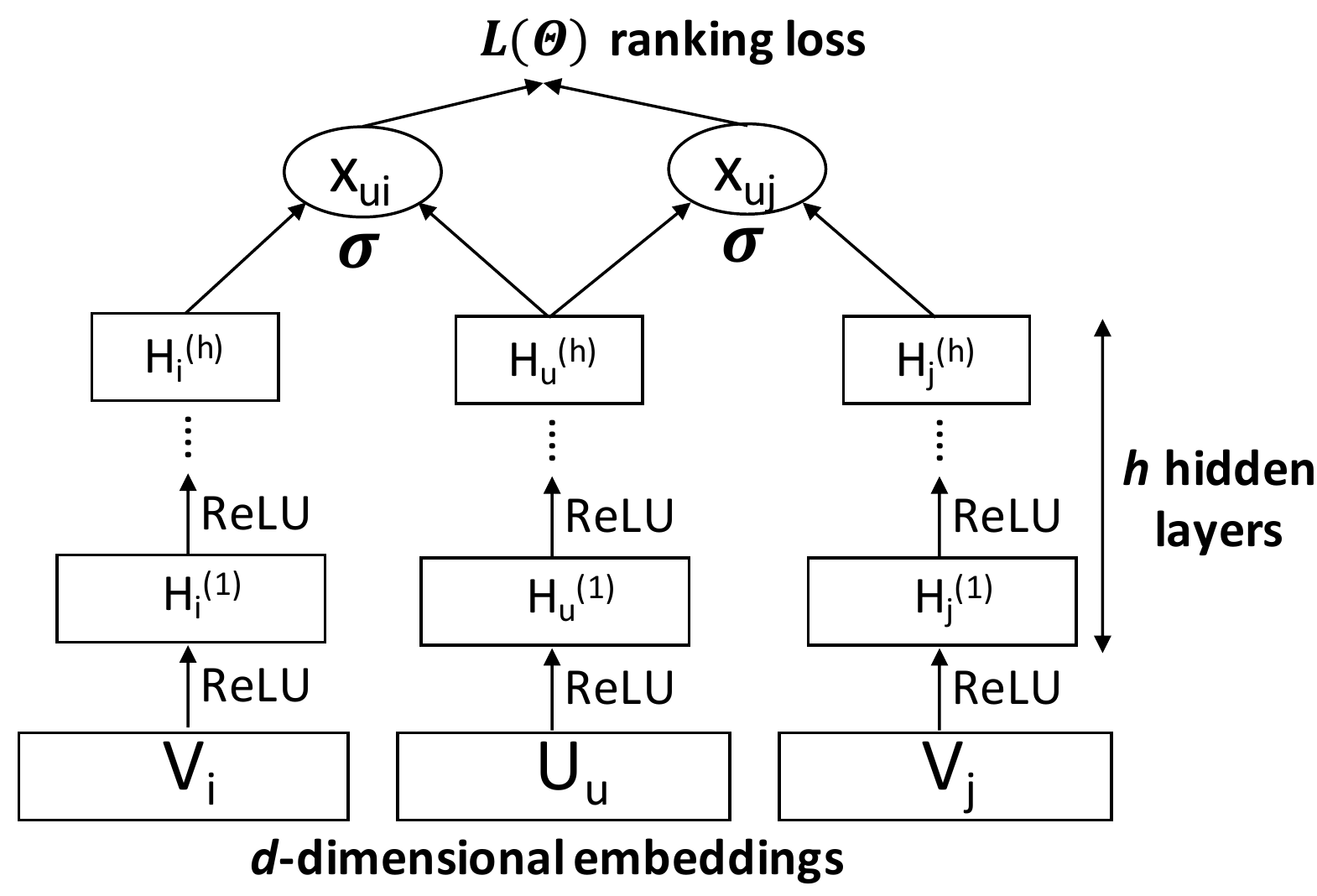}
\caption{The proposed deep pairwise learning architecture with $h$ hidden layers and the ranking loss $L(\mathbf{\Theta})$ of Eq.~(\ref{eq:loss}). For each user $u$, we consider items $i$ and $j$ based on the $r=6$ ranking criteria of Eq.~(\ref{eq:cases1}). At the bottom of our architecture, that is $H_i^{(0)}$, $H_u^{(0)}$ and $H_j^{(0)}$ we consider the respective $d$-dimensional embeddings $V_i$, $U_u$ and $V_j$. The bottom layer is the widest and each successive layer has a smaller number of hidden units. For each branch of our model we implement a tower structure, as in~\cite{p31}, halving the layer size for each successive layer.} 
\label{fig:overview}
\end{figure}

The proposed deep pairwise learning architecture is presented in Figure~\ref{fig:overview}.  We first factorize the user-item interaction matrix $X$ to compute the $d$-dimensional embeddings in the user  latent matrix $U \in \mathbb{R} ^ {n\times d}$ and item latent matrix $V \in \mathbb{R} ^ {m\times d}$, trying to minimizing the following approximation error $|| X - UV^\top ||_F^2$. Depending on the type of user feedback in the input matrix $X$ we apply different matrix factorization strategies. In case of explicit feedback e.g., ratings, we apply non-negative matrix factorization~\cite{KorenBV09}. Alternatively, in case of implicit feedback e.g., views or clicks, we follow the weighted matrix factorization strategy~\cite{p9}. Based on the $r=6$ ranking criteria of our pairwise ranking task in Eq.~(\ref{eq:cases1}), for each user $u$ we have six different sets of pairs of partial relations $(i,j) \in \mathcal{R}_u^{(r)}$. At the bottom of our architecture, that is $H_i^{(0)}$, $H_u^{(0)}$ and $H_j^{(0)}$ we consider item latent vectors $V_i \in \mathbb{R} ^ d$ and $V_j \in \mathbb{R} ^ d$, that is the $i$-th and $j$-th rows of $V$, as well as the user latent vector $U_u \in \mathbb{R} ^ d$, the $u$-th row of $U$. Then, we design the three branches of our model, where each latent vector $V_i$, $U_u$ and $V_j$ is provided to the respective branch. Given $h$ hidden layers, we first try to capture the nonlinear representations $H_i^{(q)}$, $H_u^{(q)}$ and $H_j^{(q)}$ of $V_i$, $U_u$ and $V_j$ in each branch separately, with $q=1,\dots,h$. Notice that to ease the presentation we do not show the respective user bias $b^{(q)}_u$, nor the item biases  $b^{(q)}_i$ and $b^{(q)}_j$ in the hidden layers of Figure~\ref{fig:overview}. At the top layer we calculate the probabilities $x_{ui}$ and $x_{uj}$ by combining the last hidden layers $H_i^{(h)}$, $H_u^{(h)}$ and $H_j^{(h)}$ with a sigmoid function $\sigma(x)$. To train our model via backpropagation we use the ranking loss function $L(\mathbf{\Theta})$ of Eq.~(\ref{eq:loss}) and optimize it via gradient descent. In each backpropagation step of the learning phase we use a social negative sampling strategy based on the selections of friends and foes to form the six sets of pairs of partial relations $(i,j) \in \mathcal{R}_u^{(r)}$ of Eq.~(\ref{eq:cases1}), and meet the respective six ranking criteria in the ranking loss $L(\mathbf{\Theta})$.

At this point we would like to highlight that the reason that the hidden layers of each branch of the network are learned separately, is to save the computational cost when training our model. Notice that if all hidden layers were connected, then more parameters in the weighting matrices would be required, thus significantly increasing the computational cost when optimizing the model parameters. However, the three branches are connected as follows (i) the input of the proposed model are the user latent vectors and the item latent vectors of observed and unobserved items. Notice that the latent vectors are computed by factorizing the user-item interaction matrix. The selection of unobserved items to train the parameters of the three branches of our network is performed based on our social negative sampling strategy (Section~\ref{sec:opt}). (ii) As illustrated in our neural architecture in Figure~\ref{fig:overview} the last hidden layers are combined via the sigmoid function to learn based on the joint loss function at the top of our model which is defined in Eq.~(\ref{eq:loss}) based on our ranking criteria.

\subsection{Hidden Layers} \label{sec:hid}

To optimize the minimization problem of the ranking loss function of Eq.~(\ref{eq:loss}) we compute the models' parameter set $\mathbf{\Theta}$. Given the $h$ hidden layers used in the three branches of our model of Figure~\ref{fig:overview} we set $\mathbf{\Theta}=\{W^{(q)}_i, W^{(q)}_u, W^{(q)}_j, b^{(q)}_i,  b^{(q)}_u, b^{(q)}_j\}$, $\forall q=1,\dots,h$. Matrices $W^{(q)}_i$, $W^{(q)}_u$ and $W^{(q)}_j$ are the weighting matrices of the $q$-th hidden layers to produce the deep learning representations of the latent vectors $V_i$, $U_u$ and $V_j$. Variables $b^{(q)}_i$, $b^{(q)}_u$ and $b^{(q)}_j$ denote the respective biases of the $q$-th hidden layers of each branch of our model. In our architecture the bottom layer is the widest and each successive layer has a smaller number of hidden units. This way it learns more abstract features of the $d$-dimensional embeddings and consequently better captures the nonlinear correlations between user preferences and the social information of trust and distrust relationships. For each branch of our model we implement a tower structure, as in~\cite{p31}, halving the layer size for each successive layer. Hence, to implement the tower architecture we add the constraint of $2^h \leq d$ for the number of hidden layers $h$ and the low $d$-dimensional embeddings of matrix factorization. For the hidden layers there are several choices of activation functions, like sigmoid, hyperbolic tangent  $tanh(x)$ and rectifier linear unit function $ReLU(x)$. In our implementation, we used ReLU activation functions, with $ReLU(x)= \max(0, x)$, as they are non-saturated\footnote{The saturation problems occurs when neurons stop learning and their output is near either 0 or 1, a problem that can be suffered by the sigmoid and tanh functions~\cite{p30}.}, well-suited for sparse data and making the model less likely to be overfitting~\cite{p30}. Using ReLU activation functions, $\forall q=1,\dots,h$, the $q$-th hidden layers of the three branches of our model produce the respective representations:

\begin{align}  
\begin{aligned}
& H^{(q)}_i = ReLU ( W^{(q)}_i H^{(q-1)}_i + b^{(q-1)}_i)\\
& H^{(q)}_u = ReLU ( W^{(q)}_u H^{(q-1)}_u + b^{(q-1)}_u)\\
& H^{(q)}_j = ReLU ( W^{(q)}_j H^{(q-1)}_j + b^{(q-1)}_j) 
\end{aligned}
\end{align}
with $H^{(0)}_i = V_i$,  $H^{(0)}_u = U_u$ and $H^{(0)}_j = V_j$. In addition, at the bottom layer of our architecture ($q=0$) we set the biases $b^{(0)}_i$, $b^{(0)}_u$ and $b^{(0)}_j$ to the interaction frequencies of $i$, $u$ and $j$, respectively. Notice that the user and item biases play a crucial role in recommendation systems, handling the popularity bias of users and items~\cite{KorenBV09}. For example, the song recommendation task often suffers from popularity bias. As a consequence, it will lack opportunities to discover more obscure songs and will be dominated by songs of a few popular artists~\cite{Celma08}.

In our architecture of Figure~\ref{fig:overview}, we combine the hidden representations and the biases of the last hidden layers, that is the $h$-th layers of the three branches of our model, to compute the probabilities $x_{ui}$ and $x_{uj}$ (Section~\ref{sec:prob}). We use the sigmoid function $\sigma$ to ensure that the probabilities $x_{ui}$ and $x_{uj}$ are in the range of $[0, 1]$, which are calculated as follows:
\begin{align}  
\begin{aligned}
& x_{ui}= \sigma({H^{(h)}_i}^\top H^{(h)}_u + b^{(h)}_i + b^{(h)}_u )\\
& x_{uj}= \sigma({H^{(h)}_j}^\top H^{(h)}_u  + b^{(h)}_j + b^{(h)}_u )
\end{aligned}
\end{align}

\subsection{Model Learning} \label{sec:Train} 
\subsubsection{Backpropagation and social negative sampling} \label{sec:opt}
In our implementation we used Tensorflow\footnote{\url{https://www.tensorflow.org}}, compatible with Python. We computed the model's parameter set $\mathbf{\Theta}$ via backpropagation with stochastic gradient descent, trying to solve the minimization problem of the ranking loss function $L(\mathbf{\Theta})$. The training of the network runs on CPU. We employed mini-batch Adam~\cite{p32}, which adapts the learning rate for each parameter by performing smaller updates for frequent and larger updates for infrequent parameters. We set the batch size of mini-batch Adam to 512 with a learning rate of 1e-4, and we fix the regularization parameter $\lambda$ of Eq.~(\ref{eq:loss}) to 1e-4. In each backpropagation iteration we performed negative sampling, to randomly select a subset $\mathcal{I}^-_u$ of unobserved items as negative instances. To meet the multiple ranking criteria of Eq.~(\ref{eq:cases1}), we perform \emph{social negative sampling} on the condition that (i) $\mathcal{I}^-_u \cap \mathcal{I}^+_a = \emptyset$ and (ii) $\mathcal{I}^-_u \cap \mathcal{I}^+_b = \emptyset$,  $\forall a \in \mathcal{N}^+_u$, $b \in \mathcal{N}^-_u$, that is the negative samples must not belong to the sets of observed items of all friends and foes of user $u$.  Notice that it is necessary to satisfy the two conditions of the social negative sampling strategy, as we have to produce the different sets of partial relationships based on the six cases in Eq.~(\ref{eq:cases1}). Consequently, the respective six ranking criteria can be incorporated into the ranking loss function of Eq.~(\ref{eq:loss}) over the backpropagation algorithm. In addition, in our social negative sampling strategy we draw the negative samples without replacement. In our implementation we used five negative samples for each positive/observed sample, as we found out that for larger numbers of negative samples the computational cost of the model learning did not pay off in terms of recommendation accuracy. In this setting, it took less than an hour to train the model on a single machine of 2.7 GHz Intel Core i7 with 16 GB memory. The influence of the number of negative samples on our model's performance is further studied in Section~\ref{sec:samplExp}. 

\subsubsection{Pretraining strategy} \label{sec:pretrain}
To account for the fact that the optimization method of gradient-based of Section~\ref{sec:opt} might find locally - optimal solution of the model's parameter set $\mathbf{\Theta}$, we followed a pretraining strategy~\cite{p29}.  We first trained our model with random initializations using only the user-item interaction $X$, ignoring the selections of social friends and foes in our social negative sampling strategy. In particular, in the negative sampling of the pretraining strategy we randomly select negative instances unconditionally, without considering the sets $ \mathcal{I}^+_a$ and $\mathcal{I}^+_b$. Then, we used the trained parameters as the initialization of our model with the social negative sampling strategy based on the selections of friends and foes, as discussed in Section~\ref{sec:opt}. The pretraining strategy is very important for our model. To verify this we tested our model without applying the pretraining strategy and we found that there is an average drop of -6.93\% in the model's performance. This observation has been also confirmed by relevant studies pointing out that the initialization of the model parameters $\mathbf{\Theta}$ plays a significant role for the model's convergence and performance~\cite{p26}. 

\subsubsection{Generating recommendations}
Having learned the model, that is having computed the model's parameter set $\mathbf{\Theta}$, the prediction of an unobserved item $i$ is calculated by first forwarding its low $d$-dimensional embedding $V_i$ on the respective neural network as shown in Figure~\ref{fig:overview} and then by calculating the probability $x_{ui}$. Given that the probabilities $x_{ui}$ and $x_{uj}$ are in the range of $[0, 1]$, as discussed in Section~\ref{sec:hid}, we first compute the probability of the partial relation $(i,j)$ as $(x_{ui} - x_{uj})/ 2 + 0.5$. The final top-$k$  recommendations are generated by ranking the unobserved items based on the computed probabilities. 

\section{Experiments} \label{sec:exp}

In this Section we conduct experiments aiming to answer the following research questions:

\begin{itemize}
\item[\emph{RQ1}] Compared to models that exploit trust relationships, can models with trust and distrust boost the recommendation accuracy? Does our proposed SDPL model outperform state-of-the-art models when the data scarcity of user preferences varies?

\vspace{0.1cm}

\item[\emph{RQ2}] How does our proposed social deep pairwise model perform for cold-start users with poor history record?

\vspace{0.1cm}

\item[\emph{RQ3}] Can our deep learning architecture of Figure~\ref{fig:overview} capture the nonlinear correlations between user preferences and the social information of trust and distrust relationships?

\vspace{0.1cm}

\item[\emph{RQ4}] What is the impact of our social negative sampling strategy on the proposed SDPL model?

\end{itemize}

\noindent In the following, we first present the experimental settings and enlist the compared methods, followed by answering the  four research questions.

\subsection{Experimental Settings} \label{sec:data}
In our experiments we used the Epinions dataset\footnote{\url{http://www.trustlet.org/epinions.html}}. To the best of our knowledge this is the largest publicly available dataset with user preferences and trust and distrust relationships reported in the literature~\cite{FOR15,FOR14,Rec17}. This dataset contains $n$=119,867 users, $m$=676,436 items and 12,328,927 ratings, with 452,123 trust and 92,417 distrust relationships. Notice that in the Epinions dataset trust relationships are formed if users add other users to their ``Web of Trust'', that is reviewers whose reviews and ratings they have consistently found to be valuable. In a similar spirit, distrust relationships are established if users add others to their ``Block List'', that is authors whose reviews they find consistently offensive, inaccurate or in general not valuable. Further details of how trust and distrust relationships are formed in the Epinions dataset can be found in~\cite{WWW04}. 

We train the examined models on the 50, 70 and 90\% of the ratings and evaluate them on the remaining test ratings. Instead of using the proposed social negative sampling strategy, in the pretraining strategy we use random sampling in the training sets. Having trained our model, we keep the parameters fixed to evaluate our model on the test ratings. To remove user rating bias from our results, we consider an item as relevant if a user has rated it above her average ratings and irrelevant otherwise~\cite{Rec17}. We measure the quality of the top-$k$ recommendations in terms of the ranking-based metrics recall (R@k) and Normalized Discounted Cumulative Gain (NDCG@k). Recall R@k is the ratio of the relevant items in the top-$k$ ranked list over all the relevant items for each user. NDCG measures the ranking of the relevant items in the top-$k$ list. For each user the Discounted Cumulative Gain (DCG) is defined as: $$DCG@k = \sum_{j=1}^{k}{\frac{2^{rel_j}-1}{\log_2{j+1}}}$$ where $rel_j$ represents the relevance score of item $j$, that is binary in our case, i.e., relevant or irrelevant. NDCG is the ratio of DCG/iDCG, where iDCG is the ideal DCG value given the ratings in the test set. We repeated our experiment five times, and we averaged recall and NDCG over the five runs.

\subsection{Compared Methods} \label{sec:models}

\textit{BPR~\cite{BPR}:} the baseline BPR model trying to order higher the observed items than the unobserved ones. This method samples the negative instances randomly, and does not exploit trust nor distrust relationships in the learning process.

\vspace{0.1cm}

\textit{SBPR~\cite{CIKM14}:} a ranking model that extends BPR, assuming that users tend to assign higher ranks to items that their friends prefer. SBPR exploits only trust relationships and ignores distrust ones.

\vspace{0.1cm}

\textit{SDAE~\cite{RafailidisC17}:} a deep learning strategy that couples matrix factorization with the Denoisining Autoencoders' hidden layers. SDAE uses only the selections of users' social friends.

\vspace{0.1cm}

\textit{MF-TD~\cite{FOR14}:} a matrix factorization strategy that uses a hinge loss function, so that the latent features of foes who are distrusted by a certain user have a guaranteed minimum dissimilarity gap from the worst dissimilarity of friends who are trusted by this same user.

\vspace{0.1cm}

\textit{RecSSN~\cite{TAN16}:} a recommendation method in social signed networks, that considers trust and distrust relationships when generating recommendations. RecSSN captures both local and global information from the signed graph and then factorizes both types of information with user preferences.

\vspace{0.1cm}

\textit{LTRW~\cite{Rec17}:} a ranking model that considers the relative ordering of the observed items of users, their friends and foes at the top of the recommendation list, following the optimization ranking algorithm of~\cite{WWW15}.

\vspace{0.1cm}

\textit{DPL:} a variant of the proposed SDPL model that ignores both trust and distrust relationships, by performing random negative sampling. DPL is the pretrained model of Section~\ref{sec:pretrain} used to initialize the parameters of the proposed SDPL model. The reason for using the DPL variant is to demonstrate the importance of trust and distrust relationships when they are missing in our proposed SDPL model.

\vspace{0.1cm}

\textit{SPL:} a variant of SDPL that does not perform deep learning ignoring the deep architecture of Section~\ref{sec:arch}. Instead, SPL learns the user and item latent matrices with the ranking loss function of Eq.~(\ref{eq:loss})  based on the multiple ranking criteria with trust and distrust and uses the social negative sampling strategy of Section~\ref{sec:opt}.

\vspace{0.1cm}

\textit{SDPL:} our proposed model with the social deep pairwise learning architecture of Figure~\ref{fig:overview},  that uses multiple ranking criteria in the ranking loss function based on the selections of friends and foes. 

\vspace{0.2cm}

\noindent The parameters of the examined models have been determined via cross-validation and in our experiments we report the best results. The parameter analysis of the proposed model is further studied in Sections~\ref{sec:param} and \ref{sec:samplExp}.

\subsection{Performance Evaluation - RQ1}

In Table~\ref{tab:res1}, we evaluate the performance of the examined models in terms of recall R@k and NDCG@k, with $k=10,20$. The grouped columns of Table~\ref{tab:res1} correspond to the grouping of the examined models based the source of additional information of users' social relationships employed in the learning process, as discussed in Section~\ref{sec:models}. The last group consists of the proposed SDPL model and its two variants of DPL and SPL. Next, we analyze the results reported in Table~\ref{tab:res1} for each group.

\begin{table*}[t]
\centering
\caption{Performance evaluation in terms of recall and NDCG for all users. Bold values denote the best scores, for $^*p<$0.05. The last column denotes the relative improvement (\%) when comparing SDPL with the second best method, LTRW.} 
\vspace{-0.2cm}
\begin{center}\resizebox{1.9\columnwidth}{!}{
\begin{tabular}{c@{\quad}|c@{\quad}|c@{\quad}|c@{\quad}c@{\quad}|c@{\quad}c@{\quad}c@{\quad}|c@{\quad}c@{\quad}c@{\quad}|c}\hline
&  & Baseline & \multicolumn{2}{|c|}{Models with trust} & \multicolumn{3}{|c|}{Models with trust and distrust} &  \multicolumn{3}{|c|}{Our models}  & \\ \hline
Training & Measure & BPR & SBPR & SDAE & MF-TD & RecSSN & LTRW & DPL & SPL & SDPL & Improv. (\%)\\ \hline
\multirow{4}{*}{50\%} & R@10 & 0.0487 & 0.0537 & 0.0668 & 0.0746 & 0.0763 & 0.0865 & 0.0501 & 0.0840 & \textbf{0.1016$^*$} & 17.24\\ 
& R@20 & 0.0671 & 0.0852 & 0.1034 & 0.1229 & 0.1301 & 0.1510 & 0.0723 & 0.1502 & \textbf{0.1729$^*$} & 14.48\\ 
& NDCG@10 & 0.1056 & 0.1469 & 0.1630 & 0.1734 & 0.1795 & 0.1994 & 0.1340 & 0.1911 & \textbf{0.2268$^*$} & 13.66\\ 
& NDCG@20 & 0.0644 & 0.0970 & 0.1056 & 0.1145 & 0.1202 & 0.1346 & 0.0891 & 0.1299 & \textbf{0.1531$^*$} & 13.73\\ \hline
\multirow{4}{*}{70\%} & R@10 & 0.0624 & 0.0735 & 0.0880 & 0.0991 & 0.1018 & 0.1182 & 0.0598 & 0.1112 & \textbf{0.1364$^*$} & 15.07\\ 
& R@20 & 0.0946 & 0.1247 & 0.1458 & 0.1720 & 0.1812 & 0.2146 & 0.1012 & 0.2074 & \textbf{0.2423$^*$} & 12.90\\ 
& NDCG@10 & 0.1382 & 0.1575 & 0.1756 & 0.2171 & 0.2050 & 0.2554 & 0.1384 & 0.2469 & \textbf{0.2805$^*$} & 9.83\\ 
& NDCG@20 & 0.0879 & 0.1046 & 0.1147 & 0.1456 & 0.1385 & 0.1745 & 0.0923 & 0.1698 & \textbf{0.1914$^*$} & 9.71\\ \hline
\multirow{4}{*}{90\%} & R@10 & 0.0774 & 0.0867 & 0.1099 & 0.1254 & 0.1324 & 0.1472 & 0.0801 & 0.1409 & \textbf{0.1608$^*$} & 9.23\\ 
& R@20 & 0.1246 & 0.1512 & 0.1895 & 0.2246 & 0.2465 & 0.2722 & 0.1416 & 0.2695 & \textbf{0.2915$^*$} & 7.05\\ 
& NDCG@10 & 0.1583 & 0.1856 & 0.2175 & 0.2576 & 0.2634 & 0.3116 & 0.1774 & 0.2783 & \textbf{0.3254$^*$} & 4.44\\ 
& NDCG@20 & 0.1021 & 0.1245 & 0.1447 & 0.1746 & 0.1801 & 0.2145 & 0.1201 & 0.1926 & \textbf{0.2235$^*$} & 4.14\\ \hline
\end{tabular}}\label{tab:res1}
\end{center}
\end{table*}

\vspace{0.1cm}

\textit{The baseline BPR model.} Firstly, we note that BPR has the lowest performance in the experiments for all three training sets. BPR does not utilize users' trust nor distrust relationships when producing top-$k$ recommendations. As a consequence, BPR is significantly affected by the data scarcity of user preferences in the training sets, justifying BPR's limited performance.

\vspace{0.1cm}

\textit{Models with trust.} The two trust-based models of SBPR and SDAE exploit the selections of social friends, outperforming the baseline BPR model in all runs. This occurs because both SBPR and SDAE are less affected by the scarcity in the training sets by taking into account trust relationships along with user preferences. In the deep learning strategy of SDAE, the nonlinear correlations between user preferences and trust relationships are captured, which explains the superiority of SDAE over SBPR.

\vspace{0.1cm}

\textit{Models with trust and distrust.} Next, we evaluate the performance of MF-TD, RecSSN and LTRW. Clearly, all the three models with trust and distrust beat the baseline BPR model as well as the two models with trust of SBPR and SDAE. This happens, because MF-TD, RecSSN and LTRW employ both the selections of friends and foes, which can leverage the recommendation accuracy by efficiently handling the data scarcity. The performances of the three models vary as they follow different strategies when exploiting users' trust and distrust relationships. MF-TD underperforms as it uses a pointwise loss function in its matrix factorization technique. Likewise, RecSSN calculates the correlations of users and friends/foes in the signed graph, as well as exploits the selections of users that have high reputation in the graph. However, RecSSN does not focus on the ranking performance in the top-$k$ recommendation task. As we can observe from Table~\ref{tab:res1}, LTRW outperforms both MF-TD and RecSSN. The ranking model of LTRW exploits both trust and distrust relationships, trying to balance the influence of friends' and foes' selections on user preferences at the top of the list. Notice that LTRW is the most competitive model, as it beats all the other baselines.

\vspace{0.1cm}

\textit{The DPL variant.} The deep learning strategy of the DPL variant outperforms the baseline BPR model, as DPL computes the nonlinear correlations between user preferences in the deep representations. Nonetheless, DPL does not use the multiple ranking criteria based on users' trust and distrust relationships. This means that DPL is trained in each backpropagation step based on random negative sampling. This limitation makes DPL underperform compared to the other baseline models that either consider trust or both trust and distrust relationships.

\vspace{0.1cm}

\textit{The SPL variant.} Table~\ref{tab:res1} shows that the SPL variant has comparable performance with the most competitive model of LTRW. This occurs because both models use different ranking loss functions, by incorporating the selections of friends and foes into their learning strategy to generate top-$k$ recommendations. 

\vspace{0.1cm}

\begin{table*}
\centering
\caption{Performance evaluation on cold-start users, with 70\% training set. Bold values denote the best scores, for $^*p<$0.05. The last column denotes the relative improvement (\%) when comparing SDPL with the second best method of LTRW. Compared to the performance on all users as presented in Table~\ref{tab:res1}, $\Delta$ (\%) is the relative drop of each model for the cold-start users.} 
\vspace{-0.2cm}
\begin{center}\resizebox{1.9\columnwidth}{!}{
\begin{tabular}{c@{\quad}|c@{\quad}|c@{\quad}c@{\quad}|c@{\quad}c@{\quad}c@{\quad}|c@{\quad}c@{\quad}c@{\quad}|c}\hline
& Baseline & \multicolumn{2}{|c|}{Models with trust} & \multicolumn{3}{|c|}{Models with trust and distrust} &  \multicolumn{3}{|c|}{Our models}  & \\ \hline
Measure & BPR & SBPR & SDAE & MF-TD & RecSSN & LTRW & DPL & SPL & SDPL & Improv. (\%)\\ \hline
R@10 & 0.0420 & 0.0565 & 0.0688 & 0.0829 & 0.0863 & 0.1008 & 0.0424 & 0.0951 & \textbf{0.1163$^*$} & \multirow{2}{*}{15.46}\\ 
$\Delta$ (\%) & -32.64 & -23.12 & -21.87 & -16.33 & -15.25 & -14.92 & -29.06 & -14.48 & -14.63  \\ \hline
R@20 & 0.0678 & 0.0999 & 0.1185 & 0.1473 & 0.1525 & 0.1847 & 0.0734 & 0.1782 & \textbf{0.2085$^*$} & \multirow{2}{*}{12.88}\\ 
$\Delta$ (\%) & -28.32 & -19.86 & -18.73 & -14.38 & -15.82 & -13.91 & -27.46 & -14.02 & -13.93  \\ \hline
NDCG@10 & 0.0976 & 0.1158 & 0.1317 & 0.1814 & 0.1742 & 0.2124 & 0.0991 & 0.2053 & \textbf{0.2376$^*$} & \multirow{2}{*}{11.85}\\ 
$\Delta$ (\%) & -29.38 & -26.47 & -25.04 & -16.41 & -15.03 & -16.82 & -28.44 & -16.85 & -15.29  \\ \hline
NDCG@20 & 0.0643 & 0.0834 & 0.0925 & 0.1236 & 0.1177 & 0.1495 & 0.0683 & 0.1452 & \textbf{0.1645$^*$} & \multirow{2}{*}{10.04}\\ 
$\Delta$ (\%) & -26.83 & -20.19 & -19.34 & -15.11 & -14.98 & -14.32 & -26.05 & -14.51 & -14.06  \\ \hline
\end{tabular}} \label{tab:res2}
\end{center}
\end{table*}

\textit{The proposed SDPL model.} Clearly, the proposed SDPL model performs significantly better than its two variants of DPL and SPL. The proposed SPDL model not only utilizes the multiple ranking criteria in the ranking loss function of Eq.~(\ref{eq:loss}), but also learns the nonlinear correlations of users' preference with the selections of friends and foes in its deep learning architecture at the same time. Using the paired $t$-test ($p<$ 0.05), we found out that SDPL beats the second best method of LTRW in all experiments, achieving an average relative improvement of 10.96\% for all measures. Based on the results reported in Table~\ref{tab:res1}, an interesting observation is that the relative improvement is in the ranges of $13.73-17.24\%$, $9.71-15.07\%$ and $4.14-9.23\%$ when training SDPL on 50, 70 and 90\% of the data, respectively. This observation indicates that, compared to the most competitive approach of LTRW, the proposed SDPL model achieves a larger relative improvement when a smaller training set is used. This is very important, due to the data scarcity of user preferences in the real-world setting, where models are trained in very scarce user data.

\subsection{Cold-Start Analysis - RQ2}

In the next set of experiments, we study the performance of the examined models on the cold-start scenario. We define users with less than ten appearances in the training set as cold-start users. In Table~\ref{tab:res2} we report the effect on recall and NDCG for cold-start users, having trained the examined models on the 70\% of the user data. In addition, for each measure we report the relative drop $\Delta$ (\%) of each model for the cold-start users, compared to the performance on all users as presented in Table~\ref{tab:res1}. 

We observe that for all models there is a drop on both measures in the cold-start case. In particular, models that are only trained on user preferences (BPR and DPL) have a significant drop of $\Delta$ in the range of $26.05-32.64\%$. The models with trust (SBPR and SDAE) are less influenced by cold-start users, with a drop of $\Delta$ in $18.73-26.47\%$. As shown in Table~\ref{tab:res2}, all models with trust and distrust (MF-TD, RecSSN, LTRW, SPL and SDPL) can downsize the negative effect of cold-start users on the performance, with the drop $\Delta$ being in the range of $13.91-16.41\%$. This occurs because models with trust and distrust exploit the selections of friends and foes, hence the cold-start problem has relatively less impact on the models' performances. Evaluated against the second best method of LTRW, we note that the proposed SDPL model maintains the quality of recommendations relatively high, achieving an average improvement of 14.17\% and 10.94\% in recall and NDCG, for $^*p<$0.05. This is very crucial in recommendation systems, as in real-world applications there are often many inactive or new users with poor history record, corresponding to cold-start users~\cite{KorenBV09}.

\subsection{Parameter Study of Deep Learning Architecture - RQ3 } \label{sec:param} 

\begin{figure}[t]\centering
\includegraphics[width=\columnwidth]{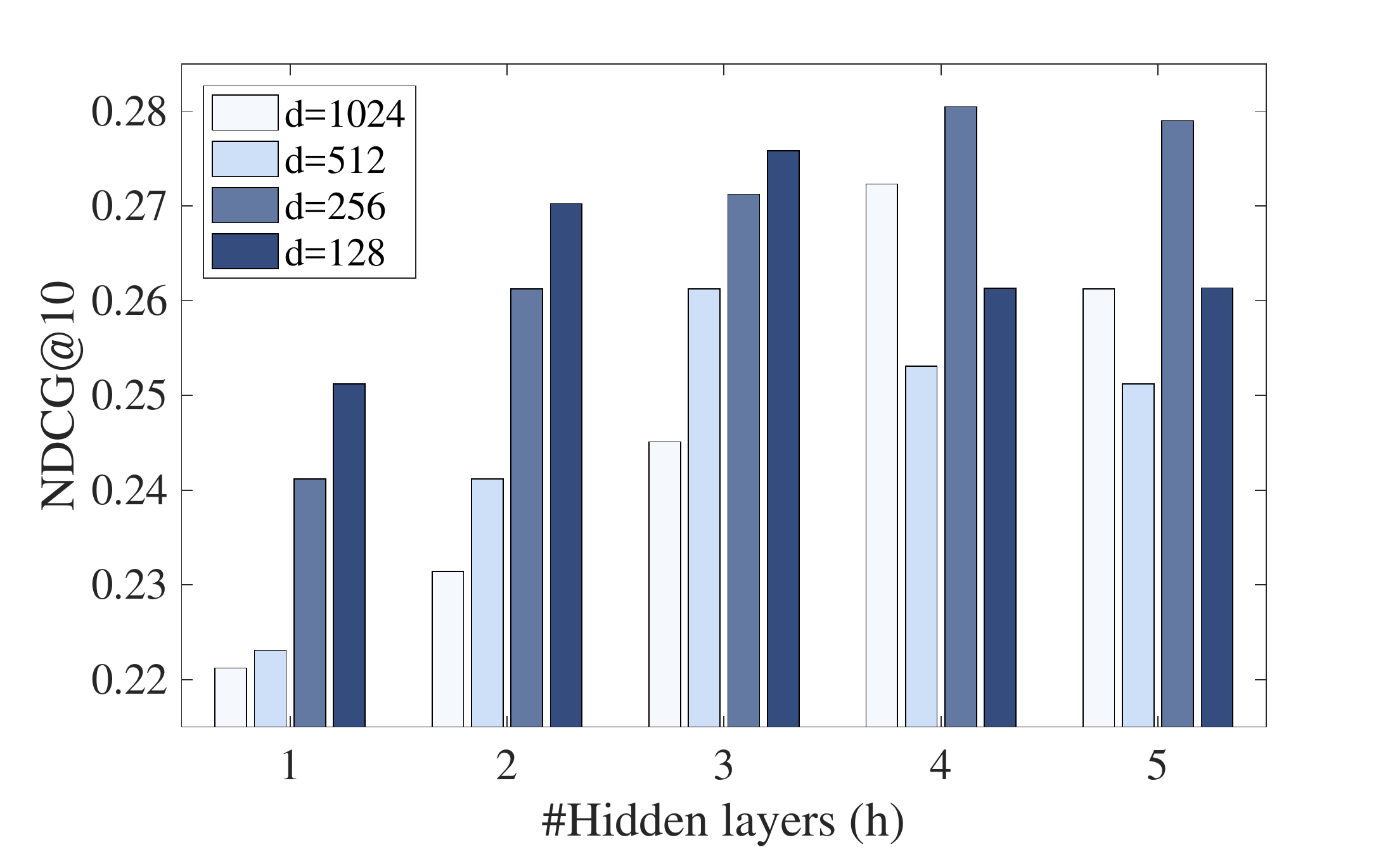}
\vspace{-0.3cm}
\caption{Impact of different deep learning architectures when varying the number of low dimensional embeddings $d$ at the embedding layer and the number of hidden layers  $h$ of the three branches of our model, subject to the constraint of $2^h \leq d$.} 
\label{fig:hid}
\end{figure}

The two most important parameters in our deep pairwise learning architecture are: (i) the number of low dimensional embeddings $d$ of matrix factorization, which are then provided to the respective branches of our model for each pairwise relation $i>_u j$, as illustrated in Figure~\ref{fig:overview}. (ii) The number of hidden layers  $h$ of the branches of our model. Given the constraint of $2^h \leq d$ of Section~\ref{sec:hid}, we vary the number of low dimensional embeddings $d$ to the power of 2. For $d=[1024, 512, 256, 128]$ we vary the number of hidden layers $h$ from 1 to 5 by a step of 1. As described in Section~\ref{sec:hid} the bottom layer ($q=0$) with the low dimensional embeddings $d$ is the widest and each successive layer has a smaller number of hidden units to learn more abstractive features of the $d$-dimensional embeddings. To better capture the nonlinear correlations between user preferences and the social information of trust and distrust relationships, we implement a tower structure for each branch of our model. This is achieved by halving the layer size for each successive layer. For example, for $d=1024$ and $h=3$ we have the following tower architecture $1024\rightarrow512\rightarrow256\rightarrow128$ or for $d=512$ and $h=2$ we have the architecture of $512\rightarrow256\rightarrow128$. 

Figure~\ref{fig:hid} shows the impact of the different deep learning architectures. We observe that the best architecture has $d=256$ and $h=4$ corresponding to the following architecture $256\rightarrow128\rightarrow64\rightarrow32\rightarrow16$. For different $d$ and $h$ values SDPL cannot capture well the nonlinear correlations between the social information of trust and distrust relationships along with user preferences, thus degrading the model's performance. In addition, we observe that SDPL becomes unstable in the cases of $d=1024$ and $d=512$ for all the different numbers of hidden layers, as well as in the case of the shallow architecture of a $h=1$ hidden layer. This observation confirms that deep learning, with $h>1$, can indeed leverage the recommendation accuracy of SDPL, capturing the nonlinear correlations between users preferences and their social relationships in the deep representations at the $h$ hidden layers.

\subsection{Social Negative Sampling - RQ4} \label{sec:samplExp}

\begin{figure}[h]\centering
\includegraphics[width=\columnwidth]{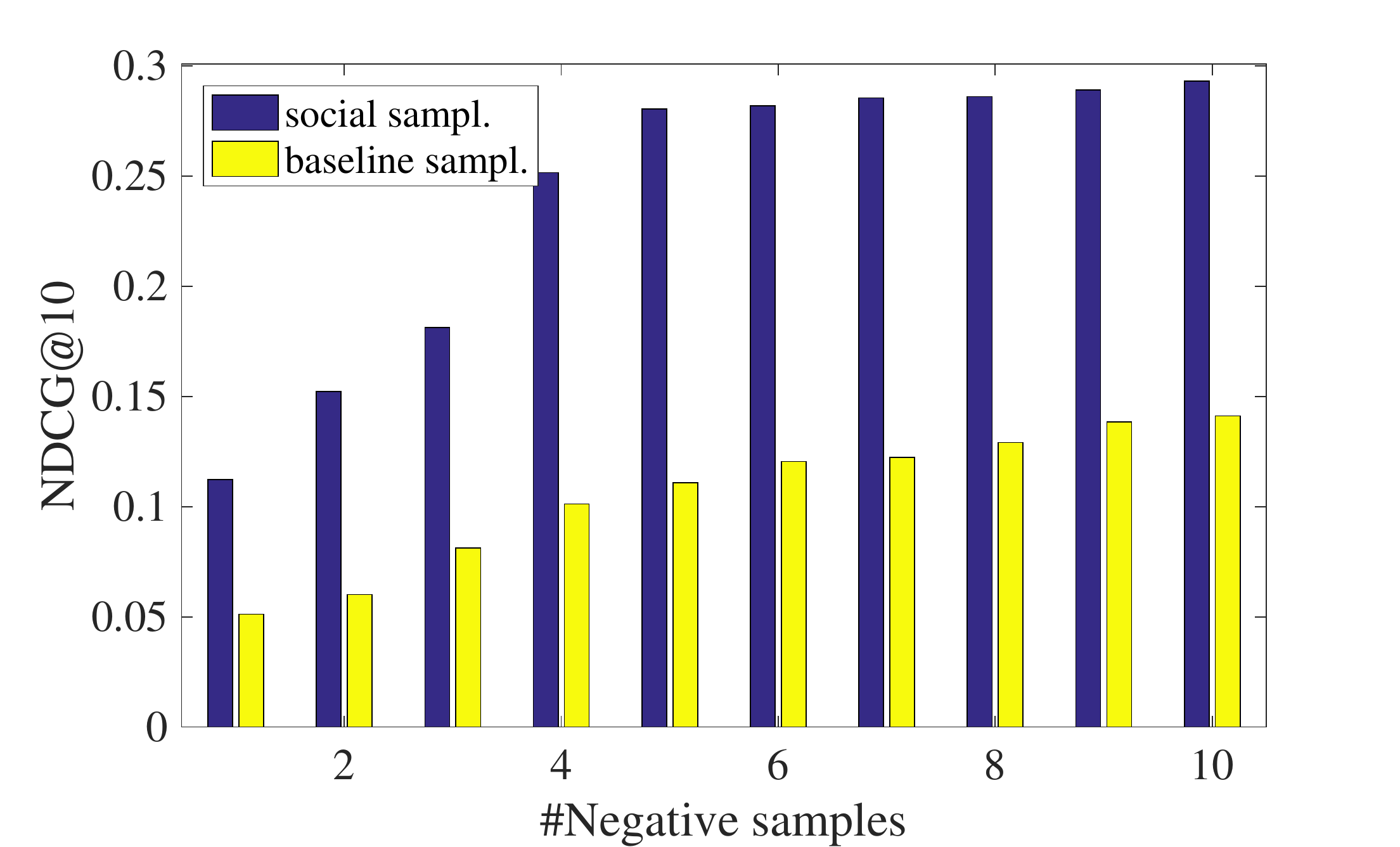}
\vspace{-0.3cm}
\caption{Effect on NDCG@10 for different numbers of negative samples when following the proposed social negative sampling strategy.} 
\label{fig:sam}
\end{figure}

In this set of experiments we evaluate the influence of the proposed social negative sampling strategy of Section~\ref{sec:opt} on the performance of the SDPL model.  Figure~\ref{fig:sam} shows the effect on NDCG@10 when we use the proposed social negative sampling strategy for different numbers of negative instances in SDPL. As reference, we also report the performance of our model when a baseline negative sampling is used. The baseline strategy ignores the selections of friends and foes when the negative samples are randomly drawn. This means that the multiple ranking criteria of the ranking loss in Eq.~{(\ref{eq:loss})} do not apply, thus the baseline strategy only uses the single set of partial relationships of Eq.~{(\ref{eq:partial})} trying to rank the observed items higher than the unobserved ones of each user $u$. 

Firstly, on inspection of Figure~\ref{fig:sam}, we note that the baseline strategy does not perform well, even in the case of drawing a large number of negative samples. This occurs because the baseline strategy aims at assigning higher ranks to the observed items of a user than the unobserved ones, without taking into account the observed items that her friends and foes prefer. Instead, the proposed social negative sampling strategy significantly boosts the recommendation accuracy, outperforming the baseline strategy in all cases. This implies that our social negative sampling strategy is more effective by selecting the unobserved items based on users' trust and distrust and consequently it produces the different sets of partial relationships based on the six cases in Eq.~(\ref{eq:cases1}). In doing so, the respective six multiple ranking criteria can be incorporated into the ranking loss function of Eq.~(\ref{eq:loss}), as discussed in Section~\ref{sec:opt}. Hence, the quality of recommendations is relatively high even for a small number of negative instances, compared to the baseline negative sampling strategy. As shown in Figure~\ref{fig:sam}, NDCG@10 does not improve much when more than five negative samples are drawn in our sampling strategy, complying also with similar observations of related studies~\cite{BPR, CIKM14}. Given that more negative samples increase the computation cost of our model, we fix the number of negative samples to five.

\section{Conclusions} \label{sec:conc}
In this paper we presented SDPL, a ranking model that performs social deep pairwise learning with users' trust and distrust relationships to face the data scarcity and the cold-start problem in recommendation systems. A key factor of our proposed SDPL model is that it aims at improving the performance in the top-$k$ recommendation task, using a \emph{ranking loss function with multiple ranking criteria} based on the selections of users and those of their friends and foes. As we experimentally showed, in comparison with the DPL variant that ignores the multiple ranking criteria, the proposed SDPL model can significantly boost the recommendation accuracy. To further verify this we compare our social negative sampling strategy with a baseline sampling strategy, and we found out that SDPL retains high recommendation quality, even when a few negative samples are drawn. Furthermore, we demonstrated the crucial role of our \emph{deep learning architecture}, as we observed that deep structures with more hidden layers are required to better capture the nonlinear correlations between user preferences and the social information of trust and distrust, compared to shallow architectures. In addition, we measure the performance of our SPL variant that ignores the deep learning architecture, and we confirmed again that the deep learning architecture of SDPL can further improve the recommendation accuracy in terms of recall and NDCG. To evaluate the performance of our SDPL model, in our experiments we varied the data scarcity of user preferences and tested our model in the case of cold-start users. Evaluated against the second best method, our SDPL model achieves an average improvement of 11.49\%  in all measures for all experiments.

Although in some web sites users can form both trust and distrust relationships, such as in  Epinions and Slashdot, in several real-world web platforms distrust cannot be computed. For example, Netflix users cannot tag others as foes nor comment on others' feedback, and as a consequence distrust relationships cannot be established. An interesting future direction is to exploit user data from various web platforms, as nowdays users open multiple accounts on different social networking websites. This is a challenging task as users behave differently in distinct social media platforms. This means that we have to weigh the knowledge transfer of user preferences along with the social information of trust and distrust relationships across the different platforms following a cross-domain strategy. In the future, to test this idea we plan to extend our SDPL model to generate recommendations for Netflix users based on user data from Epinions and Slashdot~\cite{AliannejadiRC17,RafailidisC16,RafailidisC17a,ShuWTWL18}. In addition, we plan to explore ways to extend the proposed model to account for evolving user preferences~\cite{RafailidisN14,RafailidisN16,RafailidisKM17,RafailidisN15a}, hybrid recommendations~\cite{RafailidisAEMD14} and social event detection~\cite{RafailidisSLSD13}.

\bibliographystyle{unsrt}
\bibliography{disRec}

\end{document}